\documentclass[runningheads, orivec]{llncs}

\usepackage[T1]{fontenc}
\usepackage{microtype}
\usepackage{inconsolata}
\usepackage{graphicx}
\usepackage{amsmath}
\usepackage{graphicx}
\usepackage{multirow}
\usepackage{amssymb}
\usepackage{xcolor}
\usepackage{colortbl}
\usepackage{subfigure}
\usepackage[normalem]{ulem}
\usepackage{arydshln}
\usepackage[hidelinks]{hyperref}
\usepackage{marvosym}

\def\method{HIPPO}

\begin{document}
\title{HIPPO: Enhancing the Table Understanding Capability of LLMs through Hybrid-Modal Preference Optimization}
\titlerunning{HIPPO: Hybrid-Modal Preference Optimization}
\author{
Haolan Wang\inst{1} \and
Zhenghao Liu\inst{1} \and
Xinze Li\inst{1} \and
Xiaocui Yang\inst{1} \and
Yu Gu\inst{1} \and
Yukun Yan\inst{2} \and
Qi Shi\inst{2} \and
Fangfang Li\inst{1} \textsuperscript{\Letter} \and
Chong Chen\inst{3} \and
Ge Yu\inst{1}
}
\authorrunning{Haolan Wang et al.}

\institute{School of Computer Science and Engineering, Northeastern University, Shenyang, China \\
\and 
Department of Computer Science and Technology, Tsinghua University, Beijing, China \\
\and 
Huawei Technologies Co., Ltd \\
\email{lifangfang@mail.neu.edu.cn}
}

\maketitle
\begin{abstract}
Tabular data contains rich structural semantics and plays a crucial role in organizing and manipulating information. Recent methods employ Multi-modal Large Language Models (MLLMs) to address table-related tasks across various modalities of table representations. However, existing studies mainly focus on exploring the table understanding ability of MLLMs using unimodal representations, which limits further exploration of multi-modal representations to enable more effective table reasoning. To better capture structural semantics from the tabular data, this paper introduces the \textbf{H}ybr\textbf{I}d-modal \textbf{P}reference o\textbf{P}timizati\textbf{O}n (\method{}) model, which represents tables using both text and image, optimizing MLLMs by learning more comprehensive table information from these multiple modalities. Specifically, \method{} samples MLLM responses from hybrid-modal table representations and designs a modality-consistent sampling strategy to enhance response diversity and mitigate modality bias during Direct Preference Optimization (DPO) training. Experiments on table question answering and table fact verification tasks demonstrate the effectiveness of \method{}, achieving a 4\% improvement over various table reasoning models. Further analysis reveals that \method{} not only enhances the table reasoning capability based on unimodal representations but also facilitates the extraction of complementary semantics across modalities. The code is available at \url{https://github.com/NEUIR/HIPPO}.

\end{abstract}

\section{Introduction}
\begin{figure}[t]
    \centering
    \includegraphics[width=0.95\linewidth]{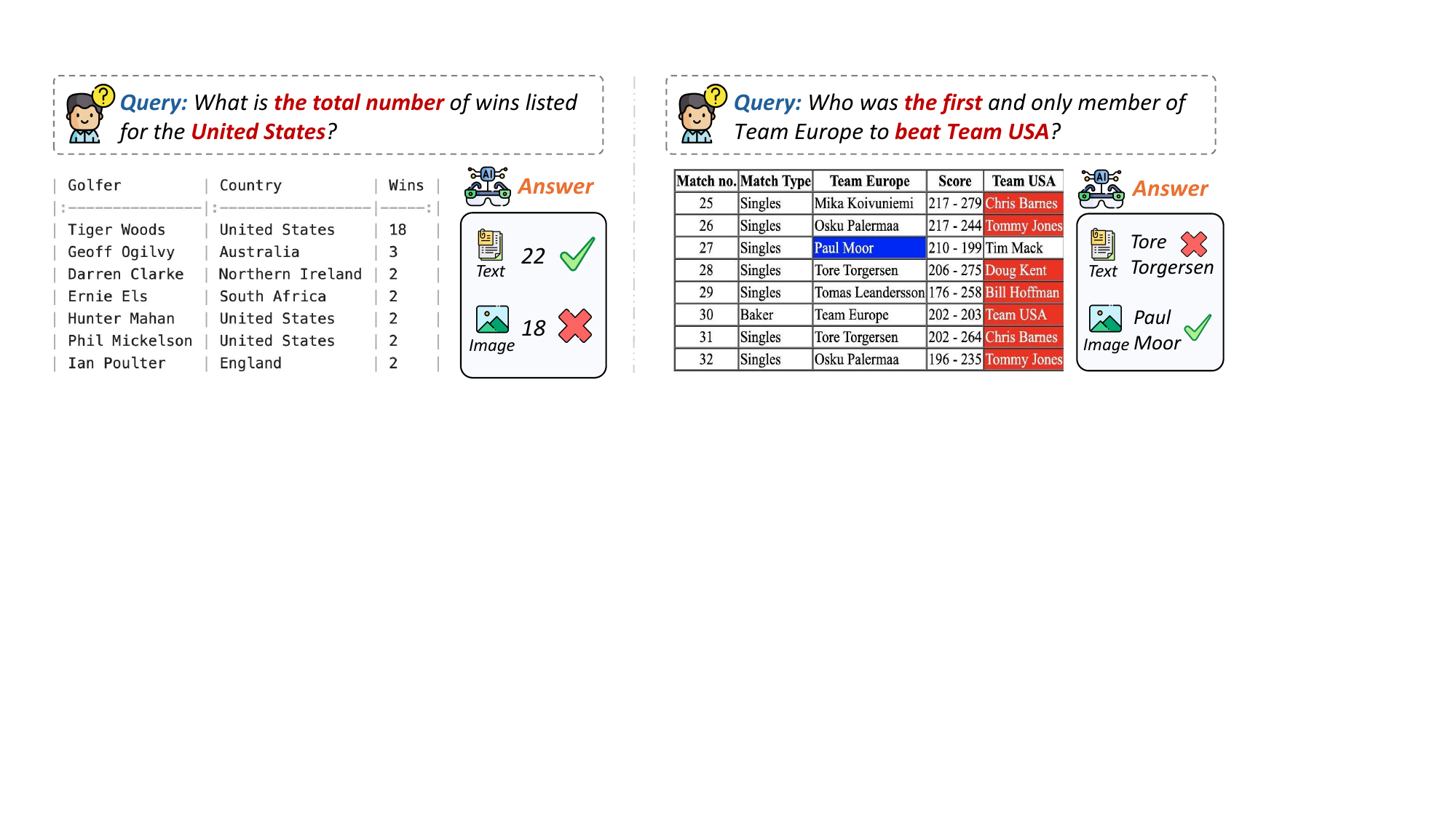}
    \caption{Performance of Both Text-Based and Image-Based Representations on the Table QA Task. We show answers generated by the MLLM (\includegraphics[width=1.1em]{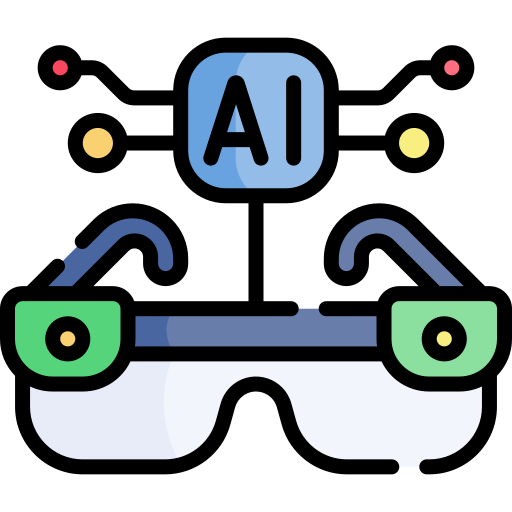}) based on both text-based (\includegraphics[width=1.1em]{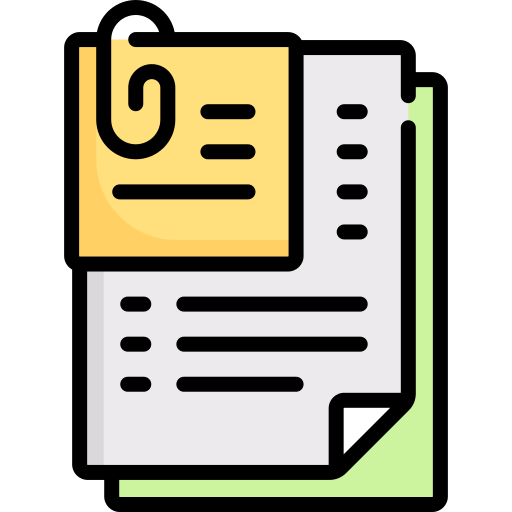}) and image-based (\includegraphics[width=1.1em]{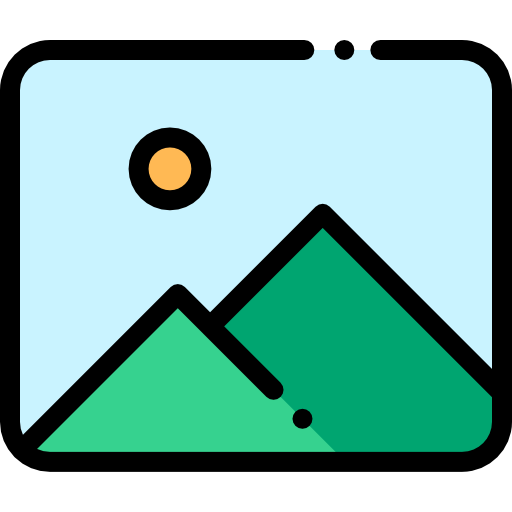}) table representations.}
    \label{fig:example}
\end{figure}

Tabular data is pervasive in daily life, widely used in databases, scientific articles, and web pages~\cite{hu2023chatdb}. The structured nature of tabular data enables the systematic organization of information into rows and columns, facilitating efficient sorting, querying, and manipulation~\cite{pujara2021tables}. Consequently, table understanding and reasoning have emerged as a significant area of interest in NLP, garnering much attention from researchers~\cite{bao2018table,zhang2024survey}.

Existing table reasoning methods primarily rely on unimodal representations. Text-based methods linearize tables and use prompts or instructions to stimulate LLMs' reasoning abilities~\cite{chen2023large,wang2024chainoftable}. In contrast, image-based methods represent tables as screenshots, leveraging MLLMs' visual perception to capture layout information such as cell alignment and background color~\cite{deng2024tables,zheng2024multimodal}.
Despite progress in unimodal scenarios, prior work rarely combines modalities to compensate for modality-specific weaknesses. Text-based representations are sensitive to formatting, and linearization often loses structural semantics, causing unstable performance across table structures~\cite{liu-etal-2024-rethinking,zhang2024flextaf,singha2023tabular}. In contrast, image-based representations preserve layout but struggle with symbolic reasoning. These complementary strengths motivate cross-modal integration for table reasoning.

As shown in Figure~\ref{fig:example}, text-based and image-based table representations each play distinct roles in enhancing the table reasoning abilities of MLLMs. Specifically, in the first case, the question asks, ``What is the total number of wins listed for the United States?'', which requires the model to identify the wins of the United States, namely ``18'', ``2'', and ``2'', and then sum them to obtain the correct answer, ``22''. The text-based table representation enables LLMs to produce the correct answer because the question relies on the arithmetic ability of language models. It facilitates precise extraction of numerical information, enabling LLMs to perform accurate calculations.
In contrast, the image-based table representation allows MLLMs to correctly answer the question in the second case. This is enabled by the visual annotation of teams with different colors to represent the win-loss situation. Both the color and cell position in the image provide crucial semantics to help MLLMs accurately answer the question. Despite these advantages, prior studies~\cite{deng2024tables,zheng2024multimodal} primarily focus on unimodal settings, leaving the potential of multi-modal representations underexplored for effective table reasoning.

To leverage the complementary strengths of text-based and image-based table representations, this paper proposes \textbf{H}ybr\textbf{I}d-modal \textbf{P}reference o\textbf{P}timizati\textbf{O}n (\method{}), a training framework that enhances MLLM table reasoning by constructing preference signals from multiple modalities. Unlike contrastive methods that align representations, \method{} uses multi-modal consistency to construct preference pairs for DPO. Specifically, \method{} prompts the same MLLM with text-only, image-only, and text+image inputs, and samples candidate responses. Instead of using randomly sampled negatives, \method{} selects a representative negative as the most frequent incorrect answer across modalities, which reflects a stable, modality-invariant error pattern.
We then perform DPO~\cite{rafailov2023direct} on the preference pairs to align the model toward the ground-truth answer while mitigating modality bias. Our main contributions are summarized as follows:
\begin{itemize}
\item We propose \method{}, a hybrid-modal preference optimization framework that jointly leverages text-based and image-based representations for robust table reasoning in MLLMs.
\item We introduce a modality-consistent negative selection strategy to obtain representative negatives via cross-modal self-consistency over text-only, image-only, and text+image inputs, enabling effective preference learning with DPO.
\item We conduct experiments and analyses on diverse table understanding benchmarks and backbones, including ablation studies and further analysis of cross-modal consistency and reasoning behavior.
\end{itemize}
\section{Related Work}

Large Language Models (LLMs), \textit{e.g.}, GPT-4~\cite{openai2023gpt} and Llama~\cite{touvron2023llama}, have demonstrated emergent abilities in table understanding through prompt and instruction methods~\cite{chen2023large,wang2024chainoftable}. Building on Chain-of-Thought (CoT) reasoning~\cite{wei2022chain}, recent studies have shown that decomposing complex tables into sub-tables can significantly improve LLM reasoning performance~\cite{zhou2022least,wang2024chainoftable,cheng2023binding}. Furthermore, some methods prompt LLMs to generate executable programs (e.g., SQL or Python) and utilize their execution results to obtain more accurate answers~\cite{cheng2023binding,ye2023large}.

Despite these advances, text-based LLMs remain sensitive to the tabular format~\cite{sui2024table,singha2023tabular}. Specifically, table understanding performance varies across different tabular formats~\cite{sui2024table}, and these formats exhibit varying levels of robustness to various noise operations~\cite{singha2023tabular}. Moreover, FLEXTAF~\cite{zhang2024flextaf} proposes an effective method to choose the most suitable text-based table representation to help LLMs answer questions. This method explores table representations using Markdown, Dict, List, Pandas, and Database formats, designing distinct mechanisms to aggregate responses across these diverse text formats.
\begin{figure*}[t]
\centering
\includegraphics[width=\linewidth]{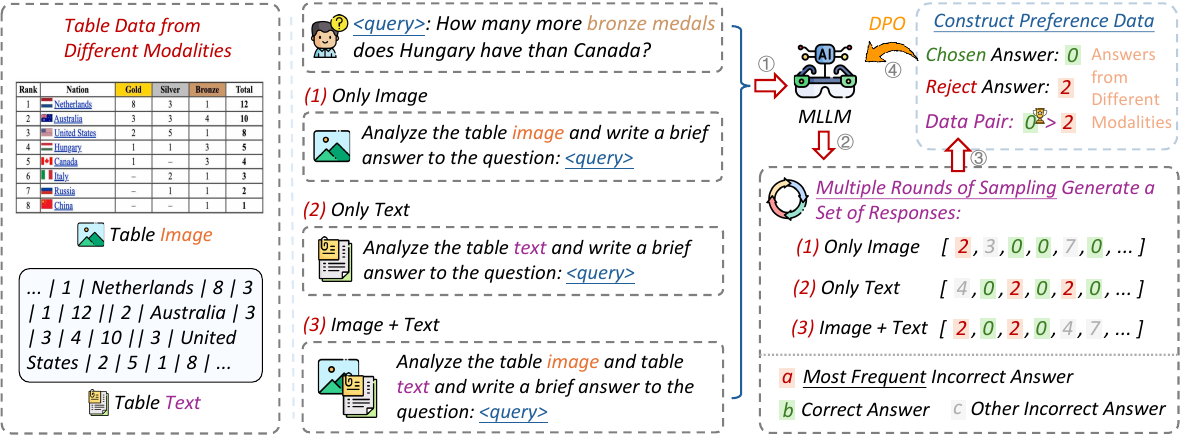}
\caption{\method{} framework. Given text and image table representations, we sample cross-modal responses under three input settings and construct DPO pairs using the ground truth as the positive and the most frequent incorrect response as the negative.} 
\label{fig:model}
\end{figure*}

With the rapid advancements in MLLMs~\cite{yao2024minicpm,bai2023qwen,liu2024improved}, recent studies have focused on image-grounded table question answering tasks~\cite{kim2024tablevqa,zheng2024multimodal}, enabling table understanding and reasoning over images from practical scenarios, such as scanned documents and web pages. Deng et al.~\cite{deng2024tables} investigate the effectiveness of text-based and image-based table representations in facilitating the table reasoning capability of LLMs and MLLMs. Their findings reveal that MLLMs exhibit robust performance with image-based table representations, and in some cases, these representations even outperform text-based ones. However, these studies have not yet explored how to effectively combine the strengths of both image and text modalities to further improve the table reasoning capabilities of MLLMs.
\section{Methodology}
As illustrated in Figure~\ref{fig:model}, this section introduces our \textbf{H}ybr\textbf{I}d-modal \textbf{P}reference o\textbf{P}timizati\textbf{O}n (\method{}) model. We begin with a detailed explanation of the multi-modal table representation (Sec.~\ref{sec:representation}). Then, \method{} leverages a hybrid-modal preference optimization approach, enabling MLLMs to effectively utilize the semantics derived from different modalities (Sec.~\ref{sec:preference}).

\subsection{Table Understanding Using Image-Based and Text-Based Representations}\label{sec:representation}
Given a table $T$ and a question $Q$, we prompt the MLLM to generate a response $y$ to answer the question $Q$ based on the table $T$. To capture both the textual semantics and the visual semantics of the table $T$, we utilize a combination of text-based and image-based table representations as inputs to the MLLM ($\mathcal{M}$), such as MiniCPM-V~\cite{yao2024minicpm}. The response $y$ is generated as follows:
\begin{equation}
y = \mathcal{M} (\text{Instruct}_\mathcal{Z}, Q, L(T), V(T)), 
\end{equation}
where $\text{Instruct}_\mathcal{Z}$ represents the instruction designed for table understanding tasks $\mathcal{Z}$, including table question answering and table fact verification. Next, we detail the construction of the text-based representation $L(T)$ and the image-based representation $V(T)$.

\textbf{Text-Based Table Representation.} To construct text-based table representations for MLLMs, existing methods typically linearize a table into its textual form, denoted as $L(T)$.

These methods utilize various data formats to convert tables into text sequences, such as Markdown, Dict, List, Pandas, and Database formats, before feeding the text-based representations into LLMs~\cite{zhang2024flextaf}. Existing works show that different table input formats will lead to different results~\cite{wang2024chainoftable,zhang2024flextaf}. To construct $L(T)$ for a given table $T$, we adopt Markdown, one of the most widely used data formats for tables.
However, text-based representations often struggle to fully capture table layout semantics. Thus, some approaches incorporate additional cell location information, such as the number of rows and columns~\cite{liu2021tapex}.

\textbf{Image-Based Table Representation.} In contrast, image-based methods directly represent a table using its screenshot, denoted as $V(T)$~\cite{zheng2024multimodal}.

A table image inherently preserves its layout, formatting, and stylistic features, providing an alternative to an intermediate textual table~\cite{sui2024table}. By leveraging the multi-modal capabilities of MLLMs, these models can efficiently perform OCR and parse table layouts, thereby enhancing document-level comprehension~\cite{luo2024layoutllm,yu2025visrag}. The image modality captures richer structural semantics, including cell positions, borders, and background colors, which significantly aid in table understanding and reasoning. Nonetheless, image-based approaches face inherent challenges in performing symbolic operations such as lookup and summation during reasoning.

\subsection{Optimizing MLLMs via Hybrid-Modal Preference Optimization}\label{sec:preference}
To enhance the table understanding ability of MLLMs, \method{} utilizes both text-based ($L(T)$) and image-based ($V(T)$) representations to enhance the semantics of the table $T$. While each modality has unique strengths and limitations, it is critical to guide MLLMs to capture more appropriate semantics from different modalities to generate accurate responses. To achieve this, \method{} proposes the Hybrid-Modal Preference Optimization method, which optimizes MLLMs using the hybrid-modal sampling-based DPO method.

\textbf{Hybrid-Modal sampling-based DPO.} The hybrid-modal sampling-based DPO method initiates by inputting both unimodal and multi-modal table representations into the MLLM ($\mathcal{M}$). For each representation, responses are sampled from $\mathcal{M}$ to construct preference pairs for DPO training:
\begin{equation}\label{eq:multimodal}
 \begin{aligned}
 \tilde{y}_l &\sim \mathcal{M} (\text{Instruct}_\mathcal{Z}, Q, L(T)),\\
 \tilde{y}_v &\sim \mathcal{M} (\text{Instruct}_\mathcal{Z}, Q, V(T)),\\
 \tilde{y} &\sim \mathcal{M} (\text{Instruct}_\mathcal{Z}, Q, L(T), V(T)).
 \end{aligned}
\end{equation}
After sampling, we collect responses from all modalities into the set $\tilde{Y} = \{\tilde{y}_l^1,\dots,\tilde{y}_l^K,\tilde{y}_v^1,\dots,\tilde{y}_v^K,\tilde{y}^1,\dots,\tilde{y}^K\}$, where $K$ denotes the number of responses sampled from each modality.

Then the positive response $\tilde{y}^+$ and the negative response $\tilde{y}^-$ are selected from the hybrid-modal candidate response set $\tilde{Y}$. The quadruples $(Q, T, \tilde{y}^+, \tilde{y}^-)$ are then collected from each table understanding task, thereby constructing the training set $\mathcal{D}$. Finally, the MLLM is optimized on the collected dataset $\mathcal{D}$ using the Direct Preference Optimization (DPO)~\cite{rafailov2023direct} method:
\begin{equation}\label{eq:dpo}
 \mathcal{L} = -\mathbb{E}_{\mathcal{D}} [\log \sigma (\beta \log \frac{\mathcal{M}(\tilde{y}^+ \mid Q, T)}{\mathcal{M}^\text{ref}(\tilde{y}^+ \mid Q, T)} 
 - \beta \log \frac{\mathcal{M}(\tilde{y}^- \mid Q, T)}{\mathcal{M}^\text{ref} (\tilde{y}^- \mid Q, T)})],
\end{equation}
where $\beta$ is a hyperparameter and $\sigma$ denotes the Sigmoid function. $\mathcal{M}^{\text{ref}}$ represents the reference model, which remains fixed throughout the training process. $\mathcal{M}$ denotes the table understanding model to be optimized during training.
Since the hybrid-modal candidate response set $\tilde{Y}$ contains responses from both unimodal and multi-modal inputs, we propose a modality-consistency based response sampling method to select more representative negative samples for DPO training.

\textbf{Modality-Consistency based Optimization.} To construct the DPO training dataset $\mathcal{D}$ for modality-consistent optimization, it is crucial to carefully select the positive response $\tilde{y}^+$ and the representative negative response $\tilde{y}^-$ from the hybrid-modal candidate response set $\tilde{Y}$. \method{} enhances the diversity of candidate responses by incorporating responses from different modalities (Eq.~\ref{eq:multimodal}). However, if the negative response is selected randomly across modalities or consistently chosen from a single modality, the model may exploit spurious modality-specific cues and develop a modality preference during optimization, as observed in multi-modal contrastive learning~\cite{liu2023universal}. Therefore, \method{} employs a modality-consistency optimization strategy that prioritizes the representative negative response that generalizes across the table representations of different modalities, by selecting the most frequent incorrect response across modalities as the negative.

Concretely, the DPO loss in Eq.~\ref{eq:dpo} is derived from the Bradley-Terry model:
\begin{equation}\label{eq:bt}
 \mathcal{L} = -\mathbb{E}_\mathcal{D} [\log \sigma (r(Q, T, \tilde{y}^+)-r(Q, T, \tilde{y}^-))],
\end{equation}
where $r(Q,T,\tilde{y})$ denotes the reward function for the generated responses. If the generated response $\tilde{y}$ matches the ground truth answer $y^*$, then $r(Q,T,\tilde{y})=1$; otherwise, $r(Q,T,\tilde{y})=0$. By minimizing the DPO loss $\mathcal{L}$, the model $\mathcal{M}$ learns to assign higher probabilities to the positive response $\tilde{y}^+$ while reducing the probabilities of the negative response $\tilde{y}^-$.

To obtain the most representative negative for optimization, thereby capturing the most common cross-modal error pattern, we construct preference pairs as follows. Specifically, we retain the query $Q$ that contains a ground truth answer within the set of model-generated responses and designate the ground truth answer $y^*$ as the positive response $\tilde{y}^+$. From the hybrid-modal candidate response set $\tilde{Y}$, we collect all incorrect responses to form $\tilde{Y}_\text{Neg}$, and select the most frequent incorrect response across modalities as the negative response $\tilde{y}^-$:
\begin{equation}
 \tilde{y}^- = \arg\max_{\tilde{y} \in \tilde{Y}_\text{Neg}} \text{Freq} (\tilde{y}),
\end{equation}
where $\text{Freq} (\tilde{y})$ denotes the occurrence frequency of the response $\tilde{y}$ in $\tilde{Y}_\text{Neg}$.
\section{Experimental Methodology}
In this section, we introduce the datasets, evaluation metrics, baselines, and implementation details.

\textbf{Datasets}. Following Zheng et al.~\cite{zheng2024multimodal}, we adopt two tasks for training and evaluation: Table Question Answering (TQA) and Table Fact Verification (TFV). The TQA task consists of five datasets, including \texttt{TABMWP}~\cite{lu2023dynamic}, \texttt{WTQ}~\cite{pasupat-liang-2015-compositional}, \texttt{HiTab}~\cite{cheng-etal-2022-hitab}, \texttt{TAT-QA}~\cite{zhu-etal-2021-tat} and \texttt{FeTaQA}~\cite{Nan2021FeTaQAFT}. For the TFV task, we use \texttt{TabFact}~\cite{2019TabFactA} and \texttt{InfoTabs}~\cite{gupta-etal-2020-infotabs}. From each dataset, we randomly sample 2,000 instances (1,800 for training and 200 for validation), except for \texttt{HiTab} and \texttt{FeTaQA}, which are used as out-of-domain test sets to assess model generalization. In total, the training set consists of 9,000 instances, and the validation set consists of 1,000 instances.

\textbf{Evaluation Metrics}. For TQA tasks, model performance is evaluated using Exact Match Accuracy (Acc.) on \texttt{WTQ}, \texttt{TABMWP}, \texttt{TAT-QA}, and \texttt{HiTab}, and BLEU score~\cite{papineni2002bleu} on \texttt{FeTaQA}. For TFV tasks, model performance is evaluated using binary classification accuracy (true/false) on \texttt{TabFact} and multi-class accuracy (entail/contradict/neutral) on \texttt{InfoTabs}.

\textbf{Baselines}. Following Zheng et al.~\cite{zheng2024multimodal}, we compare \method{} with three categories: (1) LLMs with text-based table representations, (2) MLLMs with image-based table representations, and (3) MLLMs with combined text-based and image-based table representations.

\textit{LLMs (Text)}: Tables are linearized into Markdown format for question answering.
We evaluate six LLMs: Llama2~\cite{touvron2023llama}, TableLlama~\cite{zhang-etal-2024-tablellama}, Llama3-Instruct~\cite{dubey2024llama}, Llama3.1-Instruct~\cite{dubey2024llama}, 
Qwen2.5-Instruct~\cite{yang2024qwen2}, and Qwen3~\cite{yang2025qwen3}. TableLlama is trained and evaluated on its own preprocessed table data. Since its table-processing code is unavailable, we evaluate TableLlama on both its original processed data (Oracle) and Markdown tables.

\textit{MLLMs (Image)}: Tables are provided with image-based representations to MLLMs for question answering. We compare \method{} with MiniGPT-4~\cite{zhu2024minigpt}, Qwen-VL~\cite{bai2023qwen}, mPLUG-Owl~\cite{ye2023mplug}, InternLM-XComposer~\cite{zhang2023internlm}, mPLUG-Owl2~\cite{ye2024mplug}, LLaVA v1.5~\cite{liu2024improved}, Vary-toy~\cite{wei2024small}, Monkey~\cite{li2024monkey},
TabPedia~\cite{zhao2024tabpedia}, Table-LLaVA~\cite{zheng2024multimodal}, MiniCPM-V-2.6~\cite{yao2024minicpm} and Qwen2.5-VL-Instruct~\cite{bai2025qwen2}.

\textit{MLLMs (Image \& Text)}: Tables are provided with multi-modal representations, combining both text-based and image-based inputs for answering questions. We compare \method{} with Table-LLaVA, Qwen2.5-VL-Instruct, and MiniCPM-V-2.6.

\textbf{Implementation Details}. 
For \method{} data sampling, the MLLM generates candidate responses under three table input modalities: text-based, image-based, and multi-modal. For each modality, we sample $K$ (default: $K=10$) candidate responses using a temperature of 1.0. Each response is compared with the ground truth to construct preference pairs for preference optimization training.

For \method{} training, we use the Swift~\cite{zhao2024swift} framework with a learning rate of 1e-4 and a batch size of 8, and train for two epochs using the AdamW optimizer~\cite{loshchilov2018decoupled}. To enable efficient training, we apply LoRA~\cite{hu2022lora} with configuration (rank=8, $\alpha=32$) to fine-tune the language model components. The vanilla SFT baseline is trained under the same framework and hyperparameter settings for fair comparison. During the evaluation, we use the vLLM~\cite{kwon2023efficient} framework for efficient inference.
\section{Evaluation Results}

In this section, we first present the overall performance of \method{} and perform ablation studies to evaluate the effectiveness of different training strategies. We then conduct further experiments to analyze the reasoning and modality integration abilities of \method{}. Finally, we present case studies to illustrate its reasoning process across various scenarios. 
\begin{table*}[t]
\centering
\caption{Overall Performance on Table Question Answering (TQA) and Table Fact Verification (TFV) Tasks. The \textbf{best} results are marked in bold, while the \uline{second-best} results are underlined.} \label{tab:overall}
\resizebox{\textwidth}{!}{
\begin{tabular}{l|c|ccccc|cc|c} 
\hline
\multirow{3}{*}{\textbf{Method}} & \multirow{3}{*}{\textbf{Scale}} & \multicolumn{5}{c|}{\textbf{Question Answering}} & \multicolumn{2}{c|}{\textbf{Fact Verification}} \\ \cline{3-9}
 & &  \textbf{TABMWP} & \textbf{WTQ} & \textbf{HiTab} & \textbf{TAT-QA} & \textbf{FeTaQA} & \textbf{TabFact} & \textbf{InfoTabs} & \textbf{Avg.} \\ 
 & & (Acc.) & (Acc.) & (Acc.) & (Acc.) & (BLEU) & (Acc.) & (Acc.) \\ 
\hline
\multicolumn{10}{l}{{\cellcolor[rgb]{0.957,0.957,0.957}}\textit{LLM (Text)}} \\
Llama2 & 7B & 22.82 & 16.39 & 10.72 & 13.73 & 10.93 & 9.20 & 38.92 & 17.53 \\
TableLlama + Oracle & 7B & - & 35.01 & 64.71 & - & 39.05 &  82.55 & - & 55.33  \\
TableLlama + Markdown & 7B & 10.10 & 24.97 & 46.57 & 19.04 & 38.38 & 79.37 & 46.57 & 37.86  \\
Llama3-Instruct & 8B & 42.01 & 21.24 & 6.97 & 13.08 & 12.66 & 73.89 & 54.00 & 31.98 \\
Llama3.1-Instruct & 8B & 51.86 & 26.31 & 0.08 & 10.06 & 13.60 & 77.57 & 68.44 & 35.42 \\
Qwen2.5-Instruct & 7B & 79.45 & 57.91 & 58.69 & 9.27 & 64.11 & 83.08 & 78.29 & 61.54 \\
Qwen3 & 8B & 77.49 & 72.55 & 39.21 & 4.87 & 32.38 & \textbf{91.68} & \textbf{81.62} & 57.11 \\
\multicolumn{10}{l}{{\cellcolor[rgb]{0.957,0.957,0.957}}\textit{MLLM (Image)}} \\
MiniGPT-4 & 7B & 0.22 & 0.90 & 0.20 & 0.13 & 0.39 & 0 & 0.10 & 0.28 \\
Qwen-VL & 7B & 3.30 & 0.09 & 0.06 & 0.13 & 0.45 & 1.12 & 0.65 & 0.83 \\
InternLM-XComposer& 7B & 0.06 & 0.05 & 0.12 & 0.26 & 2.62 & 1.19 & 1.11 & 0.77 \\
mPLUG-Owl & 7B & 1.76 & 0.62 & 0.25 & 0.13 & 7.42 & 7.46 & 5.53 & 3.16 \\
mPLUG-Owl2 & 7B & 6.83 & 0.67 & 0.13 & 0.39 & 11.91 & 8.21 & 26.19 & 7.76 \\
LLaVA v1.5 & 7B & 6.05 & 1.24 & 2.03 & 2.97 & 8.24 & 18.9 & 28.31 & 9.82 \\
Vary-toy & 1.8B & 4.42 & 7.96 & 3.42 & 8.81 & 2.44 & 6.33 & 6.98 & 5.77 \\
Monkey & 7B & 13.26 & 19.07 & 6.41 & 12.31 & 3.41 & 22.56 & 22.11 & 14.16 \\
TabPedia & 7B & 18.88 & 21.71 & 5.58 & 9.59 & 4.94 & 50.53 & 34.70 & 20.84 \\
Table-LLaVA  & 7B & 57.78 & 18.43 & 10.09 & 12.82 & 25.60 & 59.85 & 65.26 & 35.69 \\ 
Table-LLaVA & 13B & 59.77 & 20.41 & 10.85 & 15.67 & 28.03 & 65.00 & 66.91 & 38.09 \\
MiniCPM-V-2.6 & 8B & 83.68  & 47.97 & 56.53 & 51.55 & 32.68 & 78.48 & 72.03 & 60.42 \\
Qwen2.5-VL-Instruct & 7B & 65.63  & 56.85 & 56.82 & 50.90 & 13.92 & 81.44 & 73.46 & 56.85 \\
GPT-4o &  - & 63.27 & 56.39 & 54.82 & 25.12 & 20.76 & 83.45 & 76.00 & 54.26 \\
\multicolumn{10}{l}{{\cellcolor[rgb]{0.957,0.957,0.957}}\textit{MLLM (Image \& Text)}} \\
GPT-4o &  - & 64.14 & \uline{62.50} & 56.02 & 21.24 & 19.11 & \uline{87.62} & \uline{78.46} & 55.58 \\ \cdashline{1-10}
Table-LLaVA & 7B & 78.70 & 38.28 & 44.79 & 27.84 & \uline{34.17} & 69.42 & 63.63 & 50.98 \\
w/ \method{} & 7B & 78.22 & 41.39 & 47.01 & 43.26 & \textbf{35.37} & 74.66 & 68.71 & 55.52 \\ \cdashline{1-10}
MiniCPM-V-2.6 & 8B & \uline{86.06} & 52.30 & 58.56 & 52.46 & 32.96 & 79.31 & 73.18 &62.12 \\
w/ Vanilla SFT & 8B & 76.69 & 55.54 & \uline{62.88} & 58.91 & 16.92 & 82.54 & 76.22 & 61.93 \\
w/ \method{}  & 8B & \textbf{87.50} & 55.77 & \textbf{63.00} & 60.75 & 33.18 & 82.27 & 75.74 & \textbf{65.46} \\ \cdashline{1-10}
Qwen2.5-VL-Instruct & 7B & 62.59 & 60.45 & 61.99 & 56.73 & 11.62 & 83.43 & 75.12 & 58.58 \\
w/ Vanilla SFT & 7B & 77.17 & 61.44 & 59.49 & \textbf{62.95} & 16.79 & 80.08 & 78.13 & 62.15 \\
w/ HIPPO & 7B & 82.55 & \textbf{62.73} & 60.97 & \uline{62.82} & 10.53 & 82.51 & 78.22 & \uline{62.90} \\

\hline
\end{tabular}
}

\end{table*}

\subsection{Overall Performance}
We evaluate \method{} and baseline models on two standard tasks: Table Question Answering (TQA) and Table Fact Verification (TFV). LLMs are evaluated with textual table inputs, while MLLMs are tested with image-based or combined image-text representations.

As shown in Table~\ref{tab:overall}, LLMs with text-based inputs generally exhibit comparable performance to MLLMs with image-based inputs, highlighting the importance of both text and image modalities in table understanding. 
Among image-based MLLMs, MiniCPM-V-2.6 and Qwen2.5-VL-7B-Instruct achieve the highest overall performance across tasks.
We further incorporate text-based table representations, enabling MLLMs to perform multi-modal table understanding. The experimental results on Table-LLaVA, MiniCPM-V-2.6, and Qwen2.5-VL-7B-Instruct show performance improvements, indicating that the text representation provides complementary semantic cues to enhance the reasoning capability of MLLMs.

Subsequently, we fine-tune MiniCPM-V-2.6 and Qwen2.5-VL-7B-Instruct with text+image inputs and compare vanilla SFT with \method{} to evaluate training effectiveness. For Table-LLaVA-7B, since it has been fine-tuned on table understanding tasks, we do not report the SFT result of Table-LLaVA-7B.
Overall, \method{} outperforms both LLM (Text) and MLLM (Image) baselines on MiniCPM-V-2.6, Qwen2.5-VL-7B-Instruct, and Table-LLaVA-7B. On MiniCPM-V-2.6, \method{} outperforms the zero-shot model by 3.6\% on TQA and 2.8\% on TFV tasks, demonstrating its ability to improve multi-modal reasoning. Similarly, Qwen2.5-VL-7B-Instruct benefits from \method{}, showing overall gains on both TQA and TFV tasks. Moreover, \method{} improves the table understanding ability of Table-LLaVA-7B on TQA and TFV by 4.2\% and 5.1\%, respectively.
Furthermore, the evaluation results show that the SFT method yields inconsistent performance across different datasets and even degrades the performance of the zero-shot model on the TQA task, indicating that fine-tuning on ground-truth labels leads to overfitting. In contrast, \method{} consistently improves performance on both TQA and TFV tasks, indicating that \method{} effectively improves multi-modal table reasoning performance.

\begin{table*}[t]
  \centering
  \caption{\label{tab:ablation}Ablation on different training strategies. The best results are in \textbf{bold}, while the second-best results are \underline{underlined}. \dag and \ddag indicate statistically significant improvements over Zero-Shot and DPO Models, respectively, with $p$-value < 0.05.}

  \resizebox{\textwidth}{!}{
  \begin{tabular}{l|ccccc|cc|c}
    \hline
   \multirow{2}{*}{\textbf{Method}}  & \multicolumn{5}{c|}{\textbf{TQA}} & \multicolumn{2}{c|}{\textbf{TFV}}&  \multirow{2}{*}{\textbf{Avg.}} \\
    \cline{2-8}
    & \textbf{TABMWP} & \textbf{WTQ} & \textbf{HiTab} & \textbf{TAT-QA} &\textbf{FeTaQA} & \textbf{TabFact} & \textbf{InfoTabs} \\
    \hline
    \multicolumn{9}{l}{{\cellcolor[rgb]{0.957,0.957,0.957}}\textit{\textbf{Only Image}}} \\
    \hline
    Zero-Shot &83.68 &47.97 &\underline{56.53} &51.55 &\textbf{32.68} &78.48 &\textbf{73.03} & \underline{60.56} \\
    DPO &80.88 & \textbf{49.24} & 53.93 & 51.42 & 32.22 & 79.45 & 69.22 & 59.48\\
    \cdashline{1-9}
    \method{} (Random) &\underline{84.97}\rlap{$^{\dagger\ddagger}$}  & \underline{48.98}\rlap{$^{\dagger}$} & 54.18 & \underline{52.72} & \underline{32.64} & \underline{79.79}\rlap{$^{\dagger}$} & 70.66\rlap{$^{\ddagger}$} & \underline{60.56} \\
    \method{} &\textbf{85.71}\rlap{$^{\dagger\ddagger}$} & 48.84\rlap{$^{\dagger}$} & \textbf{57.36}\rlap{$^{\ddagger}$} & \textbf{55.95}\rlap{$^{\dagger\ddagger}$} & 32.31 & \textbf{80.14}\rlap{$^{\dagger\ddagger}$} & \underline{72.75}\rlap{$^{\ddagger}$} & \textbf{61.86}\\
    \hline
    \multicolumn{9}{l}{{\cellcolor[rgb]{0.957,0.957,0.957}}\textit{\textbf{Only Text}}} \\
    \hline
  Zero-Shot  &75.89   &50.20  &58.18  &50.90  &28.08  &74.22  &69.48 &58.14 \\
    DPO        &  65.78 & 50.96           &\underline{61.54}    &  49.74     &  27.95     & 78.91 & 68.12 & 57.57\\
    \cdashline{1-9}
    \method{} (Random) &\textbf{85.32}\rlap{$^{\dagger\ddagger}$} &\textbf{51.54}\rlap{$^{\dagger}$} & 59.51      & \underline{54.40}\rlap{$^{\dagger\ddagger}$} &    \textbf{31.40}\rlap{$^{\dagger\ddagger}$}   & \underline{79.10}\rlap{$^{\dagger}$} & \underline{74.81}\rlap{$^{\dagger\ddagger}$} & \underline{62.29}\\
    \method{} & \underline{84.71}\rlap{$^{\dagger\ddagger}$} &\underline{51.47}\rlap{$^{\dagger}$} &\textbf{62.18}\rlap{$^{\dagger}$}  & \textbf{57.38}\rlap{$^{\dagger\ddagger}$} & \underline{29.50} & \textbf{79.88}\rlap{$^{\dagger\ddagger}$} & \textbf{76.16}\rlap{$^{\dagger\ddagger}$} &\textbf{63.04} \\
    \hline
    \multicolumn{9}{l}{{\cellcolor[rgb]{0.957,0.957,0.957}}\textit{\textbf{Multi-Modality}}} \\
    \hline
    Zero-Shot &86.06 &52.30 &58.56 &52.46 &32.96 &79.31 &73.18 &62.12 \\
    DPO &83.15 &55.31 &\textbf{64.46} &58.16 &32.68 &81.38 &\underline{75.50} &64.38\\
    \cdashline{1-9}
    \method{} (Random) & \underline{86.80}\rlap{$^{\dagger\ddagger}$} &\textbf{55.93}\rlap{$^{\dagger}$} & 61.35\rlap{$^{\dagger}$} &\underline{59.32}\rlap{$^{\dagger}$} &\textbf{33.27} &\underline{81.69}\rlap{$^{\dagger}$} & 73.12 & \underline{64.50}\\
    \method{} &\textbf{87.50}\rlap{$^{\dagger\ddagger}$} &\underline{55.77}\rlap{$^{\dagger}$} &\underline{63.00}\rlap{$^{\dagger}$} &\textbf{60.75}\rlap{$^{\dagger\ddagger}$} &\underline{33.18} &\textbf{82.27}\rlap{$^{\dagger\ddagger}$} &\textbf{75.74}\rlap{$^{\dagger}$} & \textbf{65.46}\\
    \hline
  \end{tabular}
  }
\end{table*}

\begin{table}[t]
\centering
\caption{Ablation on the per-modality sampling number ($K$) for \method{} in the multi-modal setting. The best results are in \textbf{bold}, while the second-best results are \underline{underlined}.}
\resizebox{\textwidth}{!}{%
\begin{tabular}{l|ccccccc|c}
\hline
\textbf{Method} & \textbf{TABMWP} & \textbf{WTQ} & \textbf{HiTab} & \textbf{TAT-QA} & \textbf{FeTaQA} & \textbf{TabFact} & \textbf{InfoTabs} & \textbf{Avg} \\
\hline
MiniCPM-V-2.6 w/ HIPPO (k=5)  & 86.16 & 53.77 & 59.01 & 57.90 & 31.87 & 80.42 & 74.31 & 63.35 \\
MiniCPM-V-2.6 w/ HIPPO (k=10) & \textbf{87.50} & \textbf{55.77} & \textbf{63.00} & \textbf{60.75} & 33.18 & \textbf{82.27} & \textbf{75.74} & \textbf{65.46} \\
MiniCPM-V-2.6 w/ HIPPO (k=20) & 86.41 & 54.16 & 62.37 & 59.19 & \textbf{33.63} & 81.40 & 74.66 & 64.55 \\
\hline
\end{tabular}%
}
\label{tab:ablation_k}
\end{table}

\subsection{Ablation Studies}
We conduct ablation studies on MiniCPM-V-2.6 to analyze the effectiveness of \method{}. We first investigate training strategies, and then examine the per-modality sampling number $K$.

As shown in Table~\ref{tab:ablation}, we compare Zero-Shot, DPO, \method{} (Random), and \method{} on MiniCPM-V-2.6. DPO constructs preference pairs from text+image inputs with randomly chosen negatives, whereas \method{} adopts hybrid-modal sampling (text-only, image-only, and text+image) and selects negatives based on cross-modal frequency. \method{} (Random) keeps the same hybrid-modal sampling but selects negatives randomly. For fair comparison, we set $K=10$ for \method{} and \method{} (Random). In multi-modal evaluation, \method{} (Random) performs comparably to DPO, illustrating limited benefit from hybrid-modal sampling with random negatives. In contrast, \method{} achieves over 1\% improvement over the DPO method, demonstrating its effectiveness in fully utilizing the training signals to enhance the table understanding ability of MLLMs. Notably, \method{} achieves more significant improvements when feeding unimodal table representations to MLLMs, confirming the effectiveness of \method{} in capturing crucial semantics from multi-modal table representations to facilitate unimodal table understanding tasks.

As shown in Table~\ref{tab:ablation_k}, we evaluate \method{} with different per-modality sampling numbers $K\in\{5,10,20\}$. Overall performance improves when increasing $K$ from 5 to 10, suggesting that sampling more candidates per modality increases response diversity for preference construction. However, further increasing $K$ to 20 brings less consistent gains, with slight drops on most datasets, indicating that over-sampling introduces more noise and redundancy.

\subsection{Cross-Modal Consistency and Reasoning Diversity in \method{}}
In this section, we conduct a comprehensive analysis of how \method{} performs reasoning across different modalities, with MiniCPM-V-2.6 serving as the backbone model. We sample 500 examples from TAT-QA to evaluate cross-modal consistency and reasoning diversity.

\begin{figure}[t]
    \centering
    \subfigure[Jaccard Similarity.\label{fig:sim:jaccard}]{
    \includegraphics[width=0.3\linewidth]{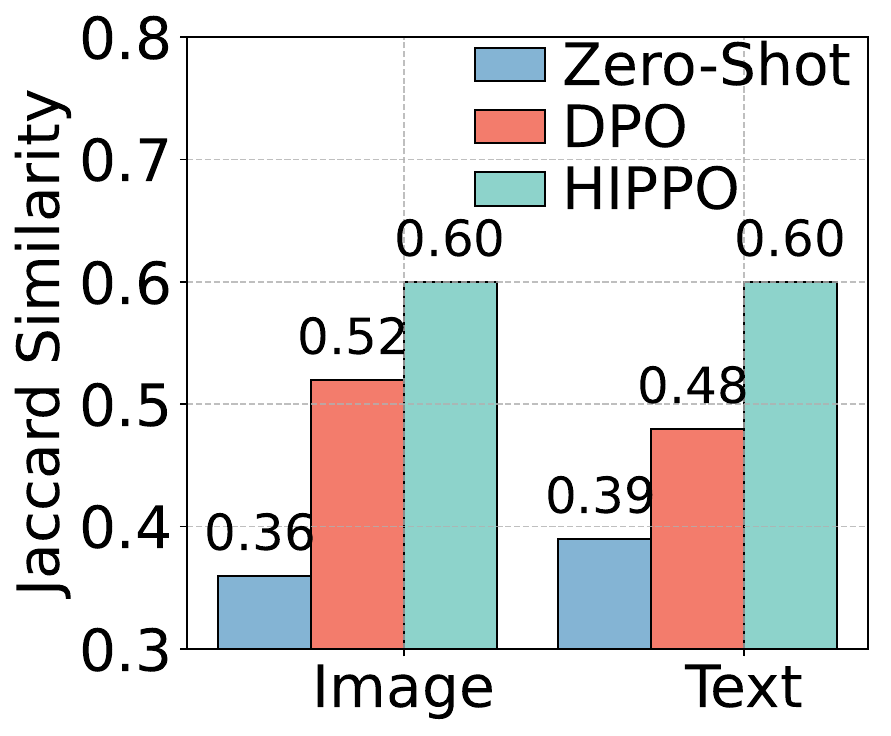}}
    \subfigure[Exact Match Acc.\label{fig:sim:acc}]{
    \includegraphics[width=0.32\linewidth]{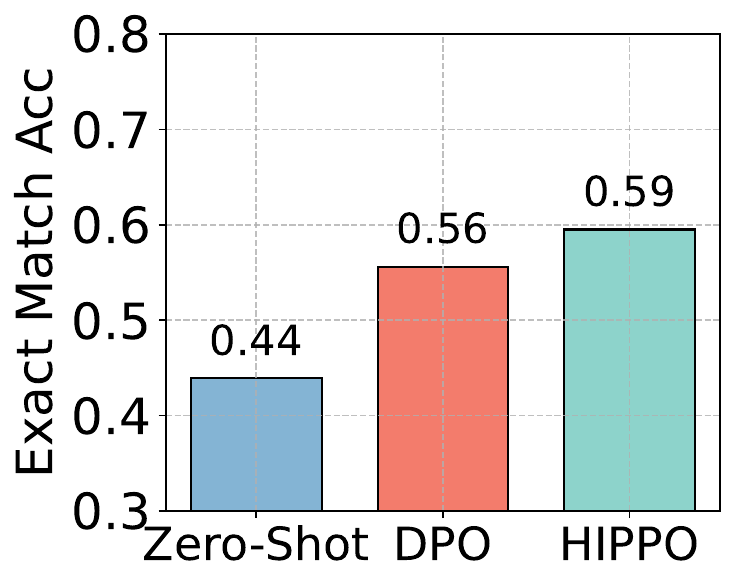}}
    \subfigure[CoT Similarity.\label{fig:sim:cot}]{
    \includegraphics[width=0.3\linewidth]{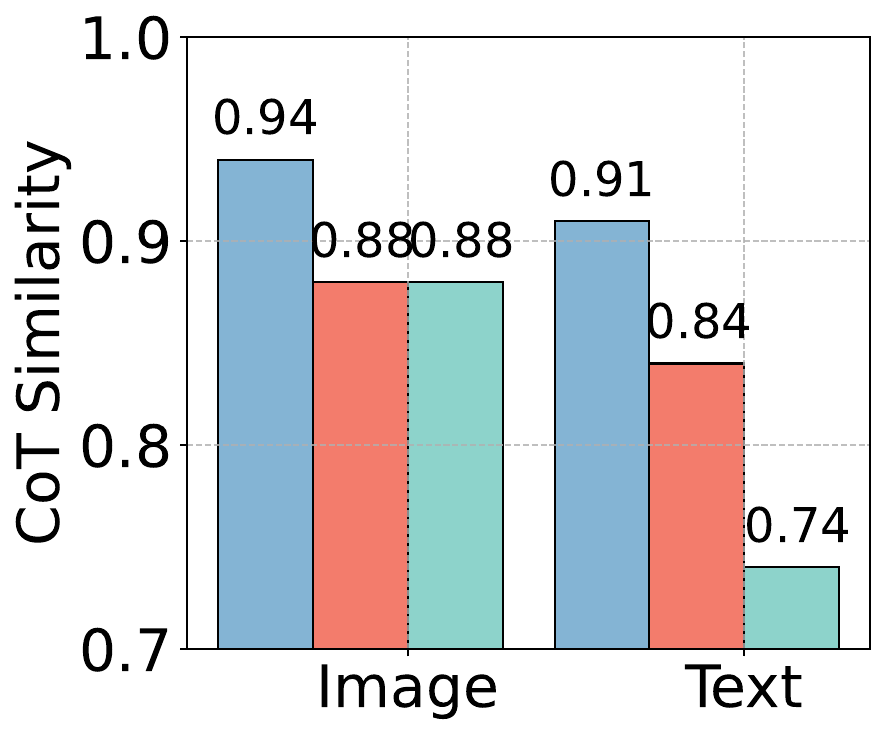}}
    \caption{Cross-Modal Consistency and Reasoning Analysis of Model Outputs under Single-Modal and Multi-Modal Table Representations.}
    \label{fig:similarity}
\end{figure}
For each question, we sample 10 responses from each model based on unimodal and multi-modal representations. We first calculate the Jaccard similarity to quantify the similarity between the outputs generated under different input conditions (unimodal vs. multi-modal). Jaccard similarity measures the ratio between the intersection and union of outputs from unimodal and multi-modal inputs, reflecting cross-modal output consistency. In addition, we evaluate answer correctness under the multi-modal setting using Exact Match Accuracy. As shown in Figure~\ref{fig:sim:jaccard}, \method{} achieves a higher Jaccard similarity score than DPO, demonstrating its effectiveness in generating more consistent responses based on both unimodal and multi-modal table representations. This suggests that \method{} enables MLLMs to better leverage semantic information without overfitting to a particular modality. Meanwhile, Figure~\ref{fig:sim:acc} shows that \method{} also attains higher Exact Match Accuracy under multi-modal inputs, indicating more accurate and reliable predictions.

To evaluate the reasoning diversity across modalities, we prompt each model to generate Chain-of-Thought (CoT)~\cite{wei2022chain} responses for answering the questions and compute the reasoning similarity using the BGE embedding model~\cite{bge}. As shown in Figure~\ref{fig:sim:cot}, CoT similarity between unimodal (image or text) and multi-modal table representations reflects the alignment of reasoning patterns across modalities. Both DPO and \method{} models exhibit reduced CoT similarity when replacing the multi-modal table representations with unimodal ones.
Compared with the DPO model, \method{} typically exhibits a lower similarity score, indicating its effectiveness in enabling MLLMs to engage in more diverse reasoning behaviors across different modalities. The reasoning diversity thrives on the modality-consistency optimization during DPO training, enabling MLLMs to better leverage information from multiple modalities.

\subsection{Multi-Modal Integration Ability under Unimodal Failures}
\begin{figure}[t]
    \centering
    \subfigure[WTQ.]{
        \includegraphics[width=0.3\linewidth]{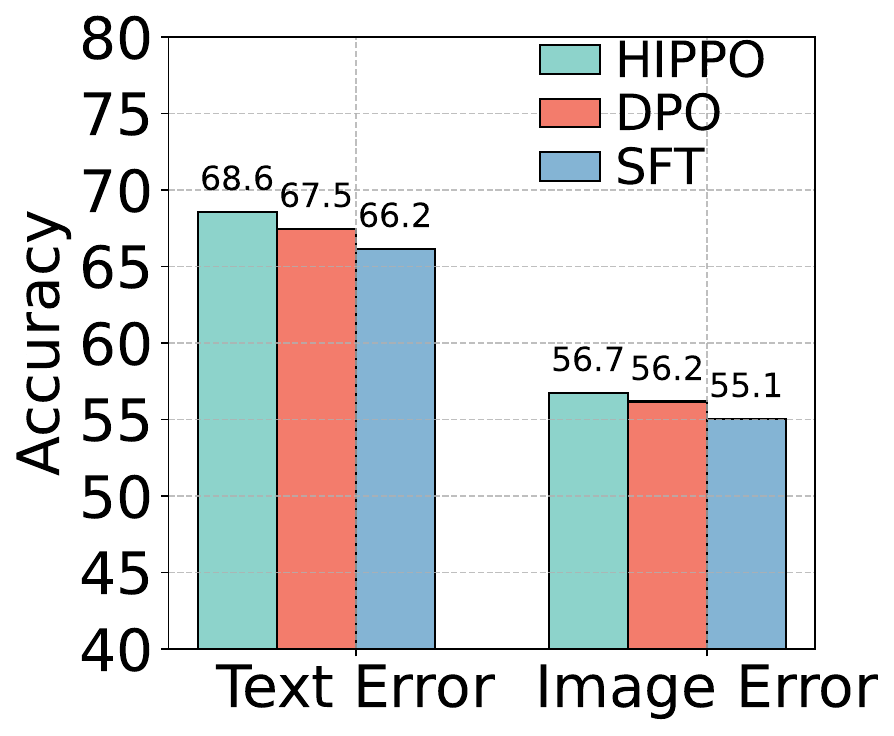} 
        \label{fig:correction_ration_wtq}
    }
    \subfigure[TAT-QA.]{
    \includegraphics[width=0.3\linewidth]{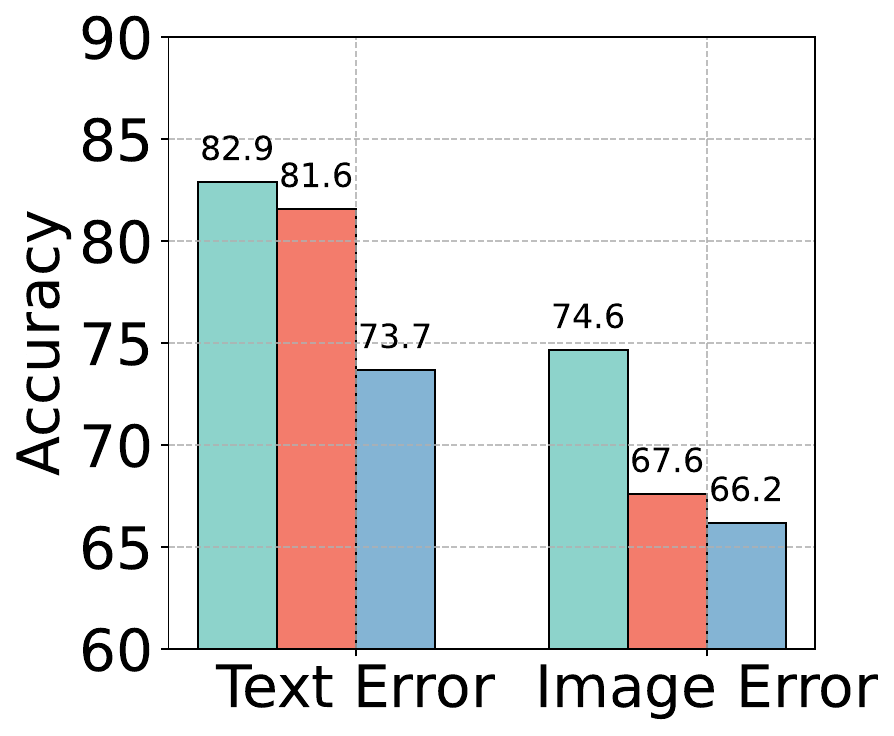}
        \label{fig:correction_ration_tat_qa}
    }
    \subfigure[TabFact.]{
        \includegraphics[width=0.3\linewidth]{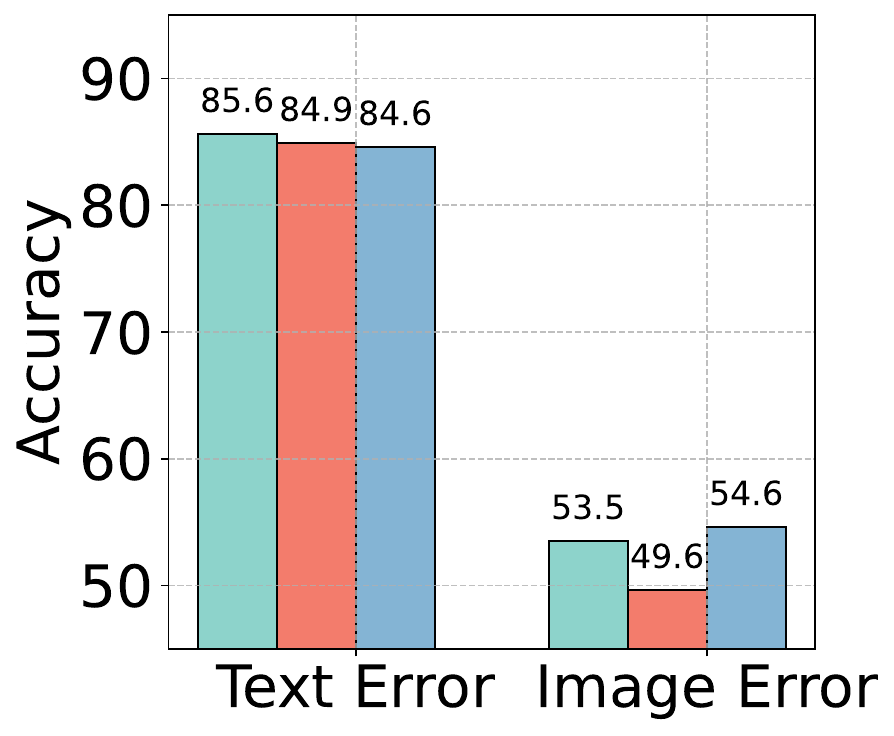}
        \label{fig:correction_ration_tabfact}
    }
    \caption{Evaluation Results of the Multi-modal Table Representations Integration Ability of Different Models.}
    \label{fig:correction}
\end{figure}
In this subsection, we evaluate \method{}'s ability to integrate information from multiple modalities under unimodal failures. The experiments are conducted with MiniCPM-V-2.6 as the backbone model. We sample 500 examples from each of the WTQ, TAT-QA and TabFact datasets.

To evaluate the integration capability of MLLMs across modalities, we construct two test scenarios: Text Error and Image Error. In the Text Error setting, we select questions that the zero-shot model answers incorrectly with text-only tables but correctly with image-only tables. In the Image Error setting, we select questions that the zero-shot model answers incorrectly with image-only tables but correctly with text-only tables. We evaluate SFT, DPO, and \method{} using multi-modal inputs on these subsets to assess how effectively each method leverages complementary cues to recover from unimodal failures.

As shown in Figure~\ref{fig:correction}, \method{} achieves the best overall performance in both the Text Error and Image Error settings, consistently outperforming SFT and DPO, benefiting from hybrid-modal training with modality-consistency-based optimization, which improves cross-modal compensation. On WTQ, all methods perform comparably under both error settings, while \method{} achieves the best results. On the more challenging dataset TAT-QA, \method{} attains the highest accuracy under both error settings. Under Text Error, DPO is only marginally behind \method{} while SFT degrades substantially; under Image Error, \method{} outperforms both DPO and SFT, suggesting more effective use of complementary cues for complex table reasoning. For fact verification dataset TabFact, all methods perform well under Text Error, but drop sharply under Image Error. DPO exhibits the largest performance drop, whereas \method{} remains robust, demonstrating the effectiveness of our modality-consistency based optimization when using visual cues of tables.

\subsection{Case Studies}
\begin{figure*}[t]
\centering
\includegraphics[width=\linewidth]{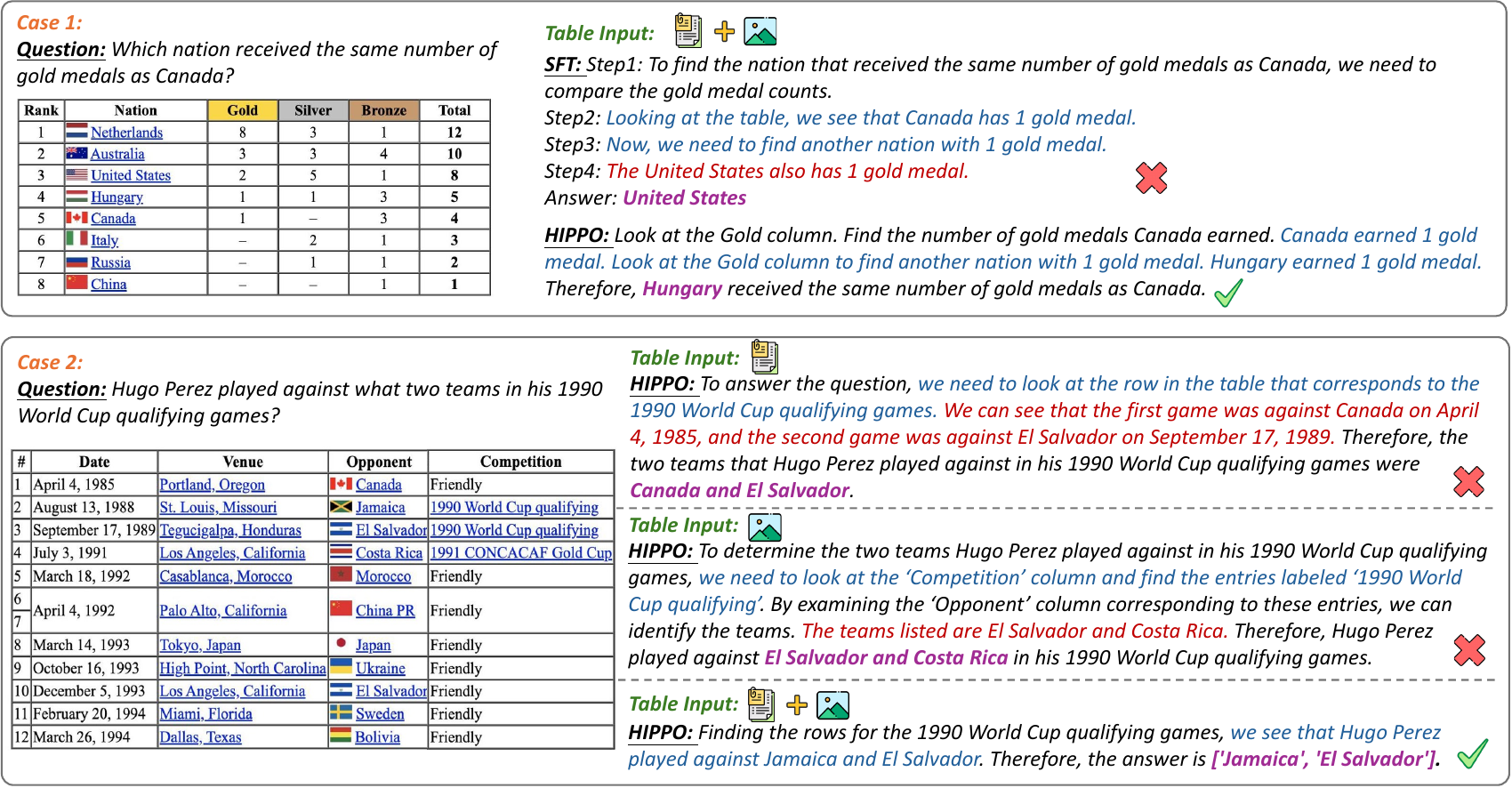}
\caption{Case Study. The \textcolor[rgb]{0.2, 0.37, 0.59}{correct reasoning}, 
\textcolor[rgb]{0.69, 0.14, 0.09}{incorrect reasoning}, and 
\textcolor[rgb]{0.58, 0.21, 0.56}{final answer} are highlighted.}
\label{fig:case_study}
\end{figure*}
As shown in Figure~\ref{fig:case_study}, we present two cases to illustrate the effectiveness of \method{}.

In the first case, the question asks, ``Which nation received the same number of gold medals as Canada?'' This question requires identifying the country that has the same number of gold medals as Canada from the table. We compare \method{} with the SFT model and show the differences in reasoning. The SFT model incorrectly states that ``The United States has 1 gold medal'', failing to make the necessary comparison across all nations to identify the correct country. In contrast, \method{} correctly identifies that ``Hungary earned 1 gold medal'', effectively integrating both textual and visual cues from the table to construct a coherent reasoning chain. By leveraging both modalities, \method{} performs a more comprehensive comparison, accurately identifying the country with the same gold-medal count as Canada.

In the second case, we feed text-based, image-based, and multi-modal table representations to the \method{} model to answer the question, ``Perez played against what teams in his 1990 World Cup qualifying games?'' This case further analyzes the role of table representations across various modalities, highlighting the impact of modality integration in complex reasoning. When given either a text-based or an image-based representation of the table, the model generates the correct intermediate reasoning step: ``find the entries labeled 1990 World Cup qualifying games'', but still incorrectly identifies ``Canada'' and ``Costa Rica'' as the teams Hugo Perez played against, revealing the inherent limitation of unimodal inputs in capturing the comprehensive contextual semantics. In contrast, with both text and image modalities, \method{} generates the correct answers, ``Jamaica'' and ``El Salvador'', demonstrating that modality integration provides complementary semantics for cross-modal validation.
\section{Conclusion}
This paper proposes \method{}, a framework designed to optimize the reasoning ability of MLLMs by integrating semantic information from multi-modal table representations, improving MLLMs table understanding capabilities. Experiment results show that \method{} achieves consistent improvements over strong LLM and MLLM baselines. Moreover, \method{} demonstrates strong generalization in unimodal settings, indicating its effectiveness beyond multi-modal inputs. Further analysis shows that \method{} enhances reasoning stability and diversity, enabling MLLMs to generate more consistent reasoning paths across modalities.
\subsubsection*{Acknowledgments}
This work is partly supported by the National Natural Science Foundation of China (No. U23B2019 and No. 62576082).
\subsubsection*{Disclosure of Interests}
The authors have no competing interests to declare that are relevant to the content of this article.

\bibliographystyle{splncs04}
\bibliography{reference}

\end{document}